\documentclass[runningheads]{llncs}
\usepackage{graphicx}
\usepackage{todonotes}
\usepackage{capt-of}
\usepackage{booktabs}
\usepackage[T1]{fontenc}
\usepackage{tabularx}
\usepackage[colorlinks]{hyperref}
\usepackage{multirow}
\usepackage{amsfonts}
\usepackage[misc]{ifsym}
\newcommand{\ie}{\textit{i.e.}}
\newcommand{\eg}{\textit{e.g.}}

\begin{document}
%



\title{Weakly-supervised Medical Image Segmentation with Gaze Annotations}

%
%

{
\hypersetup{hidelinks}

\author{Yuan Zhong$^1$ \and Chenhui Tang$^1$ \and Yumeng Yang$^2$ \and Ruoxi Qi$^2$ \and Kang Zhou$^1$ \and \\ Yuqi Gong$^1$ \and Pheng Ann Heng$^1$ \and \href{mailto:jhhsiao@ust.hk}{Janet H. Hsiao}$^{3(\textrm{\scriptsize\Letter})}$ \and \href{mailto:qidou@cuhk.edu.hk}{Qi Dou}$^{1(\textrm{\scriptsize\Letter})}$}


\authorrunning{Y. Zhong et al.}
\institute{Dept. of Computer Science and Engineering, The Chinese University of Hong Kong  \and Dept. of Psychology, The University of Hong Kong \and Division of Social Science, The Hong Kong University of Science and Technology}
%
%
\maketitle              
}
\begin{abstract}
Eye gaze that reveals human observational patterns has increasingly been incorporated into solutions for vision tasks.
Despite recent explorations on leveraging gaze to aid deep networks, few studies exploit gaze as an efficient annotation approach for medical image segmentation which typically entails heavy annotating costs.
In this paper, we propose to collect dense weak supervision for medical image segmentation with a gaze annotation scheme. 
To train with gaze, we propose a multi-level framework that trains multiple networks from discriminative human attention, simulated with a set of pseudo-masks derived by applying hierarchical thresholds on gaze heatmaps. 
Furthermore, to mitigate gaze noise, a cross-level consistency is exploited to regularize overfitting noisy labels, steering models toward clean patterns learned by peer networks. 
The proposed method is validated on two public medical datasets of polyp and prostate segmentation tasks. We contribute a high-quality gaze dataset entitled \textbf{GazeMedSeg} as an extension to the popular medical segmentation datasets. To the best of our knowledge, this is the first gaze dataset for medical image segmentation. 
Our experiments demonstrate that gaze annotation outperforms previous label-efficient annotation schemes in terms of both performance and annotation time. 
Our collected gaze data and code are available at: \url{https://github.com/med-air/GazeMedSeg}.

\end{abstract}
\section{Introduction}\label{sec:intro}
Recent studies have witnessed increasing interest in incorporating human factors into deep learning applications~\cite{li2022efficient,wang2023chatcad}. Eye tracking data, serving as a popular tool reflecting the underlying cognitive processes~\cite{yun2013exploring}, has stood out as a promising and accessible media for human-AI interaction. 
Previous works commonly utilize gaze as auxiliary information to guide deep networks~\cite{huang2021leveraging,liu2022focus,wang2023eye,wang2022follow}, with recent explorations of employing gaze as the sole supervision signal for label-efficient classification~\cite{saab2021observational}. 
However, leveraging gaze for supervising image segmentation models remains under-explored yet valuable, since it alleviates annotators' workload by alleviating the need for labor-intensive pixel-wise annotation. 
Unlike existing label-efficient annotation schemes that provide sparse supervision with bounding boxes~\cite{cheng2023boxteacher,tian2021boxinst}, points~\cite{cheng2022pointly} or scribbles~\cite{wu2023sparsely}, gaze data yield dense pixel-wise supervision signals, which is crucial for medical images featuring ambiguous boundaries and low contrast. 
These motivate us to investigate the potential of gaze-supervised medical image segmentation. 

\begin{figure}[t]
    \centering
    \includegraphics[width=1\textwidth]{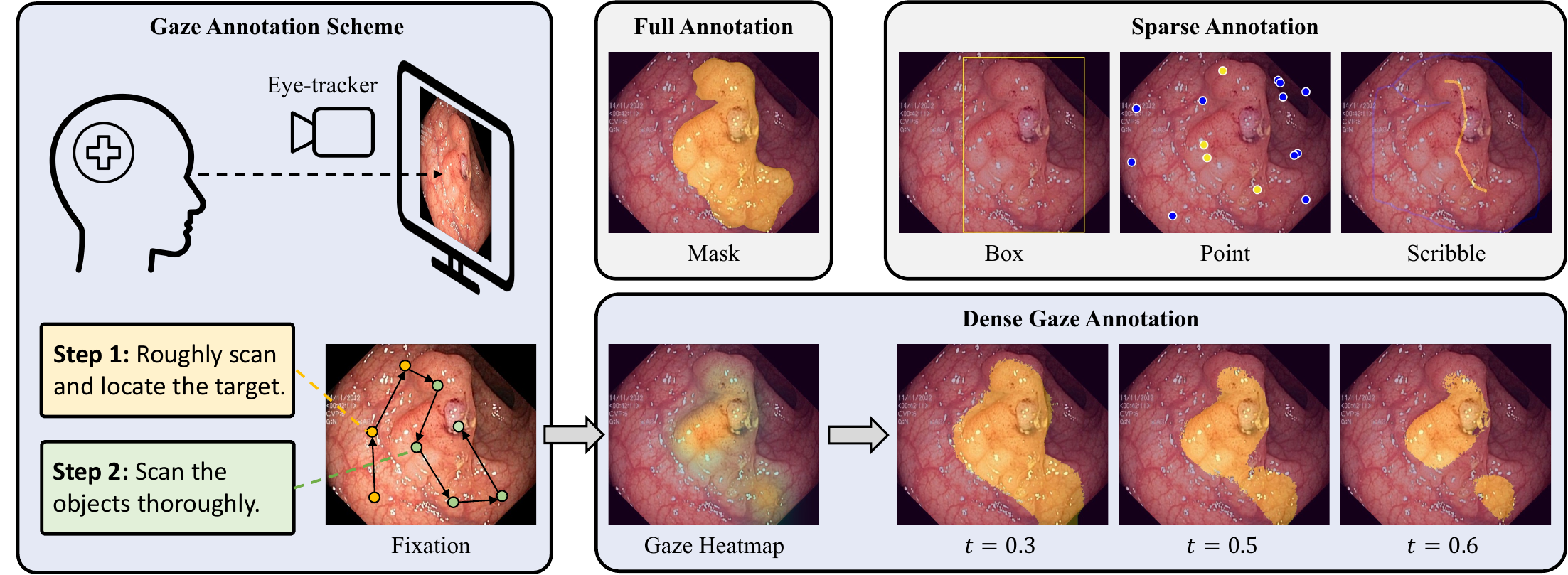}
    \caption{Illustrations of full and different label-efficient annotation schemes. Dense binarized gaze pseudo-masks are generated with various thresholds $t$, which trade off the activation of the foreground and background.}
    \label{fig:annotation_scheme}
\end{figure}

A straightforward way to gaze supervision is training with pseudo-masks generated by binarizing gaze heatmaps with a fixed threshold, where the dense gaze heatmaps contain continuous values indicating the degree of observational attention. A similar approach is widely employed in image-level semantic segmentation~\cite{chang2020mixup,wu2022adaptive} for binarizing class activation maps (CAMs)~\cite{zhou2016learning}. In practice, this approach yields suboptimal performance for gaze supervision due to the distinct characteristics of gaze data as \textit{discriminative} and \textit{noisy}.  
Firstly, the quality of pseudo-masks is sensitive to the selection of threshold (Fig.~\ref{fig:annotation_scheme}). And it is unreasonable to rive precise object boundaries with a global threshold over all images since human annotators usually pay discriminative attention even on different parts of a single object. 
Secondly, the error in eye tracking and human subjectivity makes gaze data a noisy supervision signal for segmentation. For example, the annotator may check every suspicious area when annotating the targets, thus some noisy gaze will be left. Current noise-robust approaches are based on the symmetric or asymmetric assumptions of simulated noise and design robust loss functions~\cite{han2018co,zhang2018generalized} or regularization~\cite{liu2020early}. For correlated real-world gaze noise, however, the assumptions on simulated noise do not necessarily hold.

The key to robust gaze supervision lies in the unity and consideration of the aforementioned two characteristics of gaze data. 
Inspired by multi-expert models~\cite{pavlitskaya2020using} benefiting from comprehensively integrating knowledge from multiple experts, we propose to fuse multiple diverged networks learning from multi-level human attention, simulated by applying a set of hierarchical thresholds on gaze heatmaps. These networks are designed to learn heterogeneous knowledge from discriminative human attention. 
Moreover, to mitigate gaze noise, we exploit the clean knowledge learned by peer networks of different levels to compensate for overfitting gaze noise with analysis on the memorization effect of deep networks. 

In this paper, we propose a new gaze annotation scheme that collects dense annotation in an annotator-friendly and efficient manner for segmentation tasks. Utilizing the scheme, we introduce the gaze dataset \textbf{GazeMedSeg}, which extends the Kvasir-SEG~\cite{jha2020kvasir} and NCI-ISBI~\cite{nci-isbi} datasets with gaze data of multiple annotators. 
To train with gaze, we propose a multi-level approach that trains multiple divergent deep networks to ensemble information from different levels of human discriminative attention. In addition, a cross-level consistency regularization term over predictions smoothed by a local pixel propagation module is exploited to compensate for overfitting on noisy gaze labels. 
The advantage of the proposed neat approach is in its ability to seamlessly fit into standard training pipelines with no changes to model architectures.
In experiments, we validate gaze annotation on polyp and prostate segmentation tasks using our GazeMedSeg dataset. 
Compared to the existing label-efficient annotation schemes, gaze supervision narrows the gaps with full supervision and consistently boosts performance by over 2.0\% in Dice while being 15.4\% faster to annotate, striking a sweet trade-off between performance and annotation time. 
\section{Gaze Annotation Collection}\label{sec:anno}
\subsubsection{Gaze annotation scheme.} We develop the eye-tracking program utilizing SR Research Experiment Builder platform. At the beginning of gaze annotation, each annotator goes through a 9-point gaze calibration process. 
Our gaze annotation collection consists of two stages. When presented with an image, the annotator (with eye-tracker) first roughly scans the image and locates the target objects. 
Following that, the annotator is requested to scan the objects thoroughly. 
Typically, participants start from central areas and then move on to the boundaries, ensuring that all parts of the target are covered. 
This step avoids partial activation by encouraging annotators to pay more attention to the target objects. 
Therefore, the noise will be relatively weakened when normalizing the heatmap. 
After finishing the annotating, a key is pressed to switch to the next image. 
More details on the eye-tracking settings can be found in the Appendix.

\vspace{2mm}

\noindent\textbf{GazeMedSeg dataset.}
Our collected GazeMedSeg dataset includes gaze annotations for two public medical segmentation datasets. We use the Kvasir-SEG~\cite{jha2020kvasir} dataset for polyp segmentation from gastrointestinal images, and the NCI-ISBI~\cite{nci-isbi} dataset for prostate segmentation from T2-weighted MR images. The Kvasir-SEG dataset includes 900 training and 100 testing images and the NCI-ISBI dataset includes 60 training and 10 testing volumes, where we retain slices containing prostate and obtain 789 training and 117 testing images. One annotator finishes the annotation of all images in the datasets, and we use it in our major experiments. 
We also invite two additional annotators to annotate a subset for sensitivity studies (Sec.~\ref{sec:ablation}). All annotators are experienced in medical imaging and are well-trained for eye-tracking trials. 
\section{Methodology}

\subsection{Multi-level Learning from Discriminative Attention}
The original gaze data is a series of gaze positions collected at a certain frequency. Given the gaze positions, the attention heatmaps are obtained by convolving an isotropic Gaussian over them. We further apply dense conditional random fields (CRF)~\cite{krahenbuhl2011efficient} to enhance the initial heatmaps and generate pseudo-masks. 
However, the quality of pseudo-masks varies significantly with distinct thresholds (Fig.~\ref{fig:annotation_scheme}). It is hard to balance the over-activation and under-activation of foregrounds via a fixed global threshold, because human subjectively pay discriminative attention to target objects. For instance, annotators may focus on the most discriminative part while only roughly scanning ambiguous parts of an object, which may be neglected with a fixed global threshold. 

\begin{figure}[t]
    \centering
    \includegraphics[width=1\textwidth]{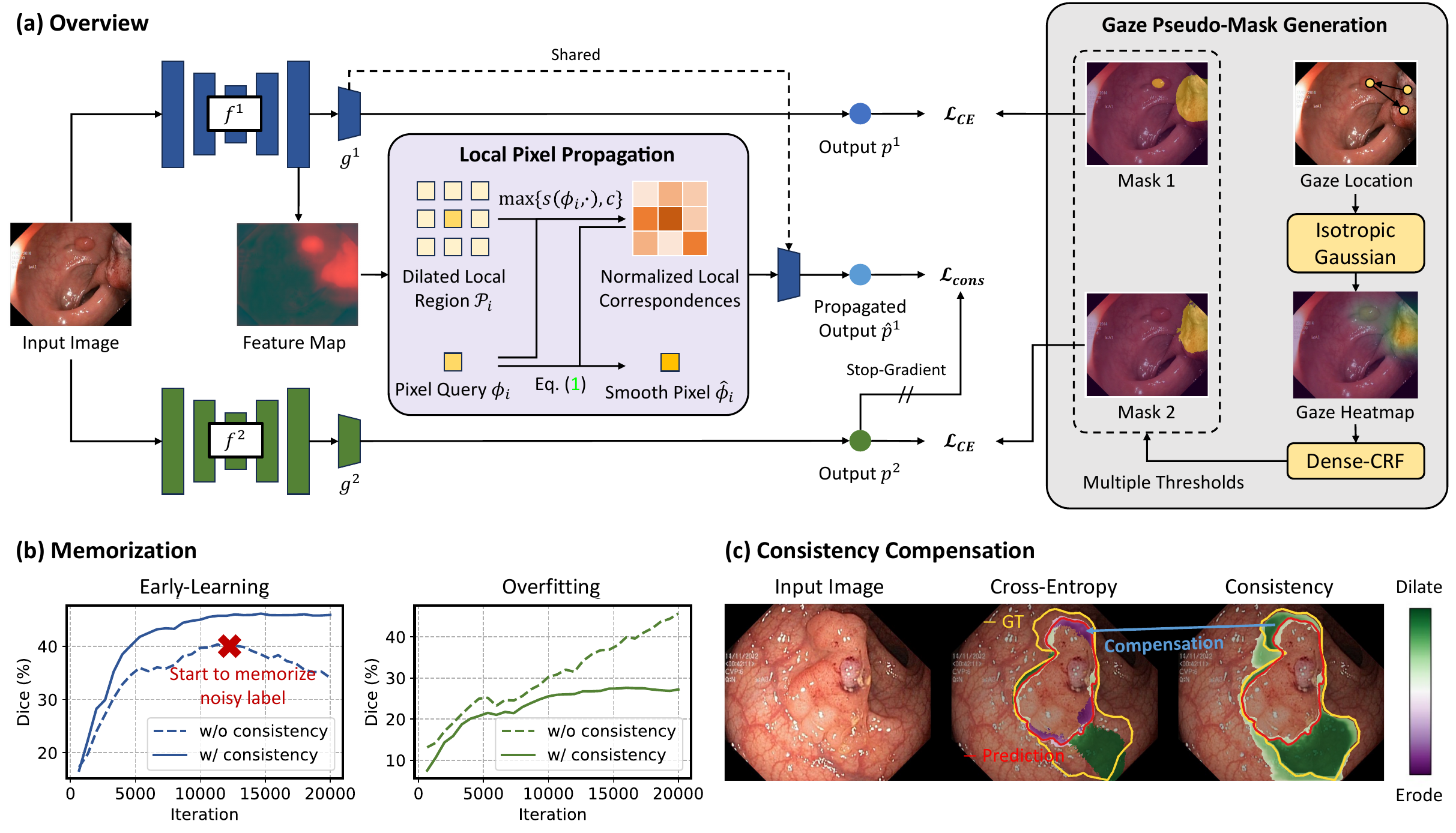}
    \caption{(a) Overview of the proposed method. For simplicity, we present the case with two levels and $\mathcal{L}_{cons}$ of network 1. The consistency loss is applied to all networks in the implementation. (b) We visualize the dynamics of \textit{early-learning} (the Dice of output and ground-truth on wrongly annotated pixels) and \textit{overfitting} (the Dice of output and noisy gaze pseudo-mask on wrongly annotated pixels) with and without the proposed consistency regularization on Kvasir-SEG~\cite{jha2020kvasir} training data. The proposed consistency prevents overfitting on the noisy labels. We use two levels and plot the average of all levels in this experiment. (c) We visualize the gradients of cross-entropy and consistency terms in the training process. The gradients that encourage dilation and erosion of the predicted target are scaled for visualization in different colors. The cross-entropy term gives noisy supervision of erosion on the top of the target object, which is compensated by consistency with clean patterns of dilation learned by other networks.}
    \label{fig:method}
\end{figure}


Our idea is to train $m$ deep networks simultaneously supervised by pseudo-masks generated from $m$ different thresholds (Fig.~\ref{fig:method}~(a)), simulating multi-level human attention. Each network is independently initialized, resulting in varying learning capabilities. Being supervised by pseudo-masks with different activation degrees, the networks evolve to learn various representations of human attention. 

In practice, we select a pair of thresholds based on annotators' feedback to generate diverse heatmaps that closely resemble ground truth and complement each other, with one tending to erode (under-activate) and the other dilate (over-activate) targets. Additional thresholds are linearly interpolated from the pair. 
Empirical results indicate that two levels are sufficient for decent performance (Sec.~\ref{sec:ablation}).
The final segmentation prediction is obtained by ensembling the predictions of these networks. We maintain the multi-level structure throughout both the training and inference stages.

\subsection{Cross-level Consistency for Noise Compensation}\label{sec:consistency}
Another essential aspect of gaze-supervised segmentation is the inevitable noise in the pseudo-masks. 
The noises can be introduced in the process of human interpretation, gaze estimation by eye-trackers, heatmap generation, etc. 
We observe a memorization effect~\cite{arpit2017closer} of deep networks when training with gaze data on the medical image segmentation task. 
As shown in Fig.~\ref{fig:method}~(b), the model captures clean patterns on incorrectly annotated pixels of pseudo-masks at the beginning of training, but eventually overfits on noisy labels. 

In the multi-level framework, though the networks are diverged with distinct supervisions, they share the same input and various pseudo-masks supervise different representations of the shared gaze data. Based on the assumption that networks learn clean patterns at the beginning of the memorization, for each network, we propose to exploit the knowledge learned by peer networks of other levels to compensate for the noisy label via a consistency term. 

To ensure noise-robust consistency, we first use a non-parametric local pixel propagation (LPP) module to 
filter the feature of each pixel by propagating the features of surrounding pixels in local regions inspired by recent works~\cite{cheng2021mitigating,hamilton2022unsupervised} proving noise-robust feature correspondence distillation. Given the feature map $\phi$, for each pixel feature $\phi_p$, the LPP module computes the transform $\widehat{\phi}_p$ as: 
\begin{equation}
\widehat{\phi}_p:=\sum_{q\in\mathcal{P}_p}\mathrm{softmax}\left(\max\{\mathrm{cos}(\phi_p,\phi_q),0\}\right)\cdot\phi_q,
\end{equation}
where $\mathrm{cos}$ denotes the cosine similarity, $\mathcal{P}_p$ denotes the set of neighboring pixels (\eg, a $3\times 3$ region with dilation 1) of pixel $p$. This refinement has a feature denoising/smoothing effect that reduces the outliers and enhances the features with local context by introducing spatial smoothness which encourages spatially close pixels to be similar. 
Given an arbitrary pair of levels $i, j\in\mathbb{Z}^+$, where $i,j\leq m$ and $i\neq j$, the consistency loss applied on the $i$-th level maximizes dot-product between the propagated prediction $\hat{p}^i$ of $i$-th level and non-propagated prediction $p^j$ of $j$-th level, \ie,   $\mathcal{L}_{\mathrm{cons}}^{(i,j)}:=-\hat{p}^i\cdot p^j$. Notably, we have $\hat{p}^i=g^i(\widehat{\phi}^i)$ for the propagated prediciton and $p^i=g^i(\phi^i)$ for the non-propagated one, where $g$ denotes the shallow classifier.


\subsection{Overall Optimization of Gaze Supervision}\label{sec:overall}
The overall loss for the $i$-th level is the combination of the supervision term and the consistency term over all other peer-level $j$:
\begin{equation}\label{eq:formulation}
\mathcal{L}^i=\mathcal{L}_{\mathrm{CE}}^i+\frac{\lambda}{m-1}\sum_{j=1,j\neq i}^m\mathcal{L}_{\mathrm{cons}}^{(i,j)},
\end{equation}
where $\mathcal{L}_{\mathrm{CE}}$ is the cross-entropy loss and can be replaced by any other segmentation loss such as dice loss, and $\lambda$ is empirically set to 3. Note that when optimizing the network of the $i$-th level, the parameters of all other networks are frozen.

The key to understanding the overall optimization process lies in the trade-off of cross-entropy and consistency terms. Intuitively, $\mathcal{L}_{\mathrm{CE}}$ trains a set of divergent networks utilizing multi-level pseudo-masks. 
However, this term tends to vanish after the early learning stage and each network starts to overfit on the respective noisy label. 
The consistency term $\mathcal{L}_{\mathrm{cons}}$ compensates for it, implicitly forcing networks to continue learning from clean patterns learned by networks of other levels. 
The mechanism can be viewed as pushing networks to struggle to find a balance between divergence and consistency, in which the hyper-parameter $\lambda$ controls this balance. It is worth noting that both divergence and consistency are equally essential for optimization. 
The consistency term ensures robustness to noise and the cross-entropy term expands and diversifies clean knowledge learned and prevents collapsing into a single network.
\section{Experiments}
We validate the proposed gaze annotation scheme on the aforementioned Kvasir-SEG~\cite{jha2020kvasir} dataset for polyp segmentation and NCI-ISBI~\cite{nci-isbi} dataset for prostate segmentation. 
For all datasets, we only utilize weak annotations for training and report performance on the testing set. 
We train a 2D UNet~\cite{ronneberger2015u} from scratch for 15k iterations with a NVIDIA A40 GPU. 
We use Adam optimizer with batch size 8 and learning rate $4e^{-4}$. More details on the training recipe can be found in the Appendix. 

\subsection{Comparison Among Label-efficient Annotation Schemes}
\subsubsection{Comparison with state-of-the-art weakly-supervised methods.} \label{sec:sota}
\begin{figure}[t!]
 \begin{minipage}{.59\linewidth}
 \centering
 \captionof{table}{Comparison with full mask supervision and SOTA weakly-supervised methods using different annotation schemes. We report the mean and standard deviation of three runs with different seeds. Dice is used as the evaluation metric. The reported annotation time is estimated to annotate 900 images in Kvasir-SEG~\cite{jha2020kvasir} training set.}
 \vspace{10pt}
 \resizebox{1\textwidth}{!}{
 {\renewcommand{\arraystretch}{1.2}
    \begin{tabular}{@{\hspace{3pt}}c@{\hspace{3pt}}|@{\hspace{3pt}}c@{\hspace{3pt}}|c@{\hspace{4pt}}|@{\hspace{3pt}}c@{\hspace{3pt}}|@{\hspace{3pt}}c@{\hspace{3pt}}}
    \hline\hline
         \multirow{2}{*}{Method} & \multirow{2}{*}{Sup.} & \multicolumn{2}{c|@{\hspace{3pt}}}{Polyp} & Prostate \\
         \cline{3-5}`
         && $\ $Anno. Time & Dice & Dice\\
         \hline\hline
         Vanilla & Full & 18.7 hrs & $82.12_{\pm 1.11}$ & $80.58_{\pm 0.48}$\\
         \hline
         BoxInst~\cite{tian2021boxinst} & Box & 3.1 hrs & $65.72_{\pm 2.97}$ & $73.78_{\pm 1.15}$ \\
         BoxTeacher~\cite{cheng2023boxteacher} & Box & 3.1 hrs & $73.33_{\pm 1.30}$ & $75.60_{\pm 1.15}$ \\
         PointSup~\cite{cheng2022pointly} & Point & 4.8 hrs & $73.05_{\pm 1.64}$ & $73.46_{\pm 4.71}$ \\
         AGMM~\cite{wu2023sparsely} & Point & 4.8 hrs & $75.57_{\pm 0.84}$ & $73.86_{\pm 1.26}$ \\
         AGMM~\cite{wu2023sparsely} & Scribble & 2.6 hrs & $67.23_{\pm 1.02}$ & $72.70_{\pm 1.03}$  \\
         Ours & Gaze & \textbf{2.2 hrs} & $\mathbf{77.80_{\pm 1.02}}$ & $\mathbf{77.64_{\pm 0.57}}$ \\
    \hline\hline
    \end{tabular}}
    }
    \label{tab:main}
\end{minipage}\quad
\begin{minipage}{.385\linewidth}
\centering
    \includegraphics[width=1\textwidth,height=0.75\textwidth]{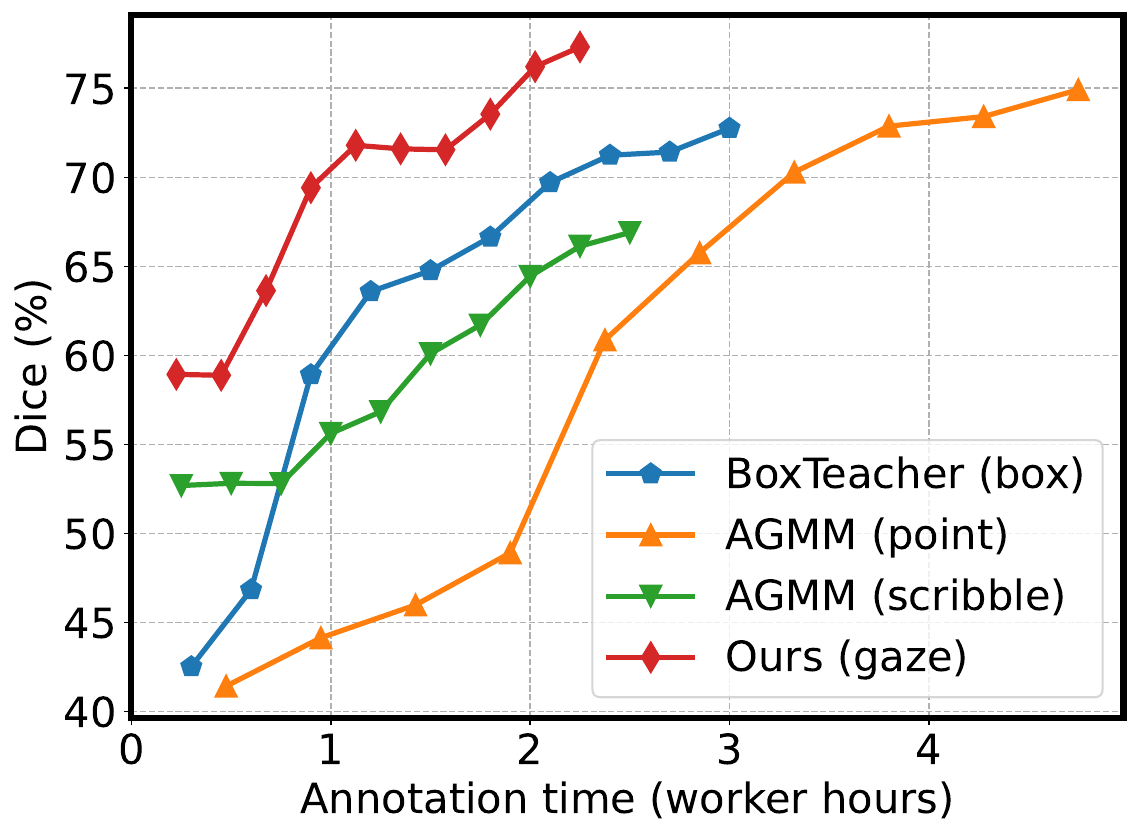}
    \caption{Performance versus annotation time for different annotation schemes. To match annotation times among annotatíion forms, we train a 2D UNet model using from 10\% to 100\% of the Kvasir-SEG training set.}
    \label{fig:tradeoff}
\end{minipage}
\end{figure}

We compare the new gaze-based annotation scheme with full mask supervision and other state-of-the-art weakly-supervised methods using different sparse annotations including box, point, and scribble on two datasets. 
To compare annotation time, we also invite the same annotator to annotate a randomly sampled subset of the Kvasir-SEG dataset containing 100 images using other annotation schemes and report the annotation time in Table~\ref{tab:main}. 
Note that we annotate the bounding box using extreme points clicking~\cite{papadopoulos2017extreme}, and annotate points using the scheme suggested by~\cite{cheng2022pointly}.
The estimated time is close to that reported in the literature~\cite{cheng2022pointly,kuznetsova2020open,papadopoulos2017extreme,valvano2021learning}, where the narrow gaps may come from the difference in the complexity of different target objects to be annotated.
For all datasets, we simulate weak annotations based on the ground truth. Note that we randomly sampled 10 pixels and 10 background pixels inside and outside the bounding box respectively as suggested by~\cite{cheng2022pointly} for point annotations, and we follow~\cite{valvano2021learning} to simulate scribble annotations. In Table~\ref{tab:main}, our results show that gaze supervision outperforms previous weakly-supervised methods trained with other sparse annotation schemes and achieves over 95\% of the fully-supervised performance.





\begin{figure}[t]
    \centering
    \includegraphics[width=1\textwidth]{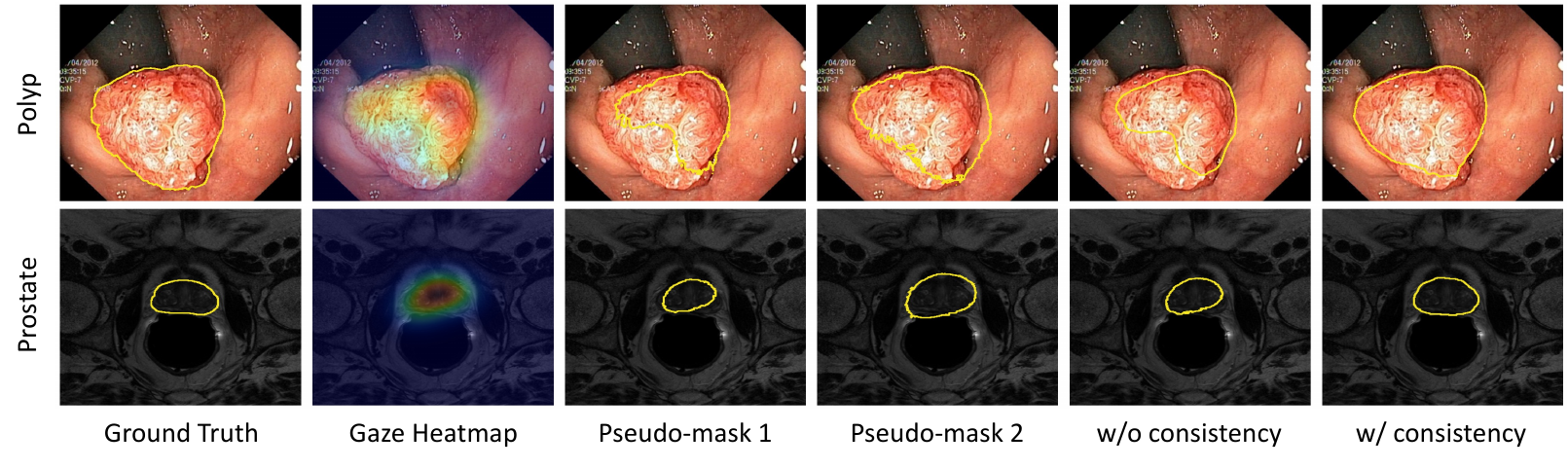}
    \caption{Visualization of gaze data and predictions. The model without consistency term ensemble the noise of different levels. Instead, the model regularized by consistency learns clean patterns of pseudo-masks and demonstrates robustness to noises.}
    \label{fig:visualization}
\end{figure}

\vspace{2mm}

\noindent\textbf{Trade-off between performance and annotation time.}
We compare the proposed gaze annotation scheme with other label-efficient sparse annotation schemes for image segmentation under the same annotation budget, \ie, the time required to annotate training data. Fig.~\ref{fig:tradeoff} presents our results on Kvasir-SEG~\cite{jha2020kvasir}, proving that gaze annotation boosts weakly-supervised segmentation by striking a sweet performance/annotation time trade-off by maximizing performance with the least annotation time among existing weak annotation schemes. 

\begin{figure}[t]
    \centering
    \includegraphics[width=1\textwidth]{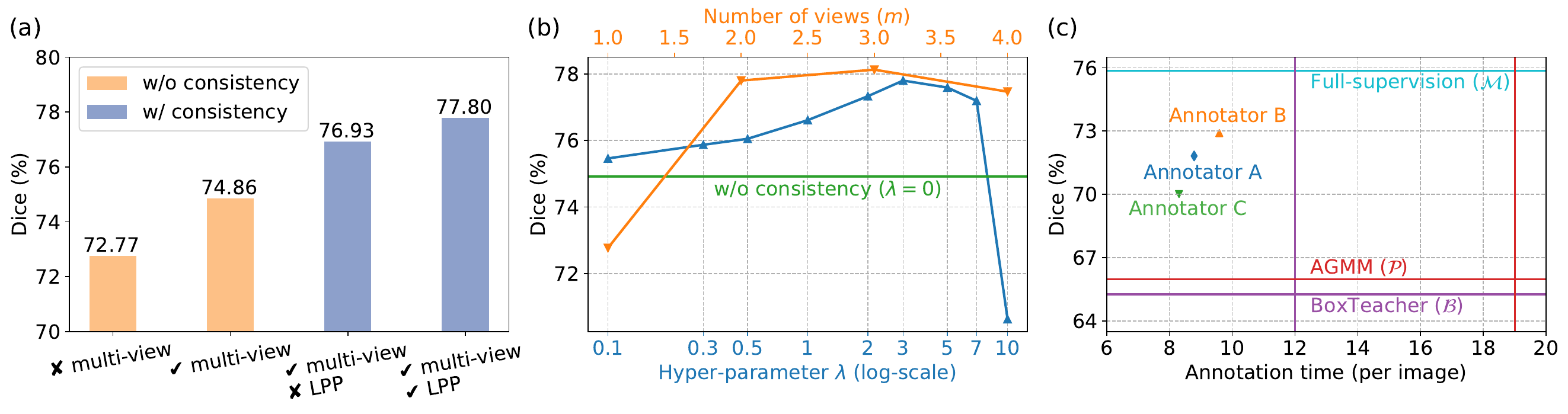}
    \caption{Ablation studies on polyp segmentation. (a) Effects of proposed components. (b) Effects of the hyper-parameter $m$ ($m=2$ by default) and $\lambda$ ($\lambda=3$ by default). The orange and blue lines deficit the results of varying $m$ and $\lambda$, respectively. (c) Effects of three different annotators (major annotator A and additional annotators B and C) and comparison with other annotation schemes.}
    \label{fig:ablation}
\end{figure}

\subsection{Ablation Studies}\label{sec:ablation}
\subsubsection{Modules and hyper-parameters.} To evaluate the impact of critical components of the proposed method, we study the effectiveness of the proposed components for gaze supervision in Fig.~\ref{fig:ablation}~(a). 
We further present the result of different choices of hyper-parameter $m$ (number of levels) and $\lambda$ in Fig.~\ref{fig:ablation}~(b). Our experiments show that $m=2$ is optimal for gaze training while increasing $m$ gives limited benefits but leads to greater training and inference complexity. The results on $\lambda$ echo the intuition of it in Sec.~\ref{sec:overall}. We also observe that having $\lambda$ greater than 7 results in a degenerated model that collapses to consistency with performance worse than simply ensembling without consistency regularization. We further visualize gaze-supervised predictions in Fig.~\ref{fig:visualization}, where the model with consistency regularization demonstrates resistence to gaze noise.

\vspace{2mm}

\noindent\textbf{Sensitivity to annotator.}
We invite two additional annotators to annotate a subset of the Kvasir-SEG training set containing 500 images and train a UNet on this subset using different annotation schemes. Note that all annotators receive the same training for gaze annotating. The results presented in Fig.~\ref{fig:ablation}~(c) show that though eye-tracking is subjective, different annotators demonstrate comparable annotation time and supervision quality, consistently outperforming other annotation schemes.
\section{Conclusion and Future Work}\label{sec:con}
In this paper, we propose to train deep networks with gaze annotations efficiently collected using a new gaze annotation scheme for medical image segmentation. The proposed method can be seamlessly integrated into standard training pipelines. The results show that gaze annotation achieves a sweet performance and annotation time trade-off compared to other annotation forms.

Our explorations give rise to several potential directions for future work: (1) While we expect gaze supervision to be broadly applicable to medical applications, the multiplied complexities in the training and especially the inference stage hamper real-time scenarios since multiple networks are maintained even though we have shown that $m=2$ is sufficient. One potential direction is to aggregate networks at a certain frequency in the training and only keep an aggregated model for inference. (2) Though specialized hardware is required to collect gaze data, we foresee that eye-tracking will not pose a bottleneck for clinical practicality even at present with the advance of commercial VR/XR headsets with precise and affordable eye-tracking capabilities. (3) We focus on binary segmentation in this paper, and the extension to multiple cases is straightforward via annotating each class separately and deciding the label for each pixel as the class with the highest value in gaze heatmaps of different classes.

\vspace{2mm}
{
\small
\noindent\textbf{Acknowledgments}. This work was supported by Hong Kong Research Grants Council (Project No. T45-401/22-N), and Science, Technology and Innovation Commission of Shenzhen Municipality (Project No. SGDX20220530111201008).
}

\vspace{2mm}
{
\small
\noindent\textbf{Disclosure of Interests}. The authors have no competing interests to declare that are relevant to the content of this article. 
}

%
%
\bibliographystyle{splncs04}
\bibliography{Paper-1675}

\clearpage
\section*{A. Gaze Data Collection}

\begin{table}[h!]
    \centering
    \caption{Eye tracking settings to build the GazeMedSeg dataset using SR Research Experiment Builder. Major settings for the program building are reported.}
    {\renewcommand{\arraystretch}{1.2}\begin{tabular}{l@{\hspace{8pt}}|@{\hspace{8pt}}c@{\hspace{8pt}}}
\hline\hline

Eye Tracker & SR Research EyeLink 1000 Plus\\
Tracked Eye & Monocular (ocular dominance)\\
Sampling Rate & 1000 Hz \\
Tracking Error & $\leq 0.5^{\circ}$ visual angle\\
\hline
Screen-Eye Distance & $46-55$ cm \\
Screen Size & 21.5 inch \\
Screen Resolution & $1024\times 768$\\
Screen Refresh Rate & 60 Hz \\
Displayed Image Size & $768\times 768$\\
Displayed Image Position & Centered \\
Displayed Image Extent & Vertical \& Horizontal: $\approx 54$ cm\\
\hline
Using Chin Rest & Yes \\
Wearing Glasses & No (contact lenses are allowed) \\
\hline
Tracking Program & SR Research Experiment Builder\\
Host PC OS & Microsoft Windows 10 \\

\hline\hline
\end{tabular}}
\end{table}

\begin{figure}[h!]
    \centering
    \includegraphics[width=1\textwidth]{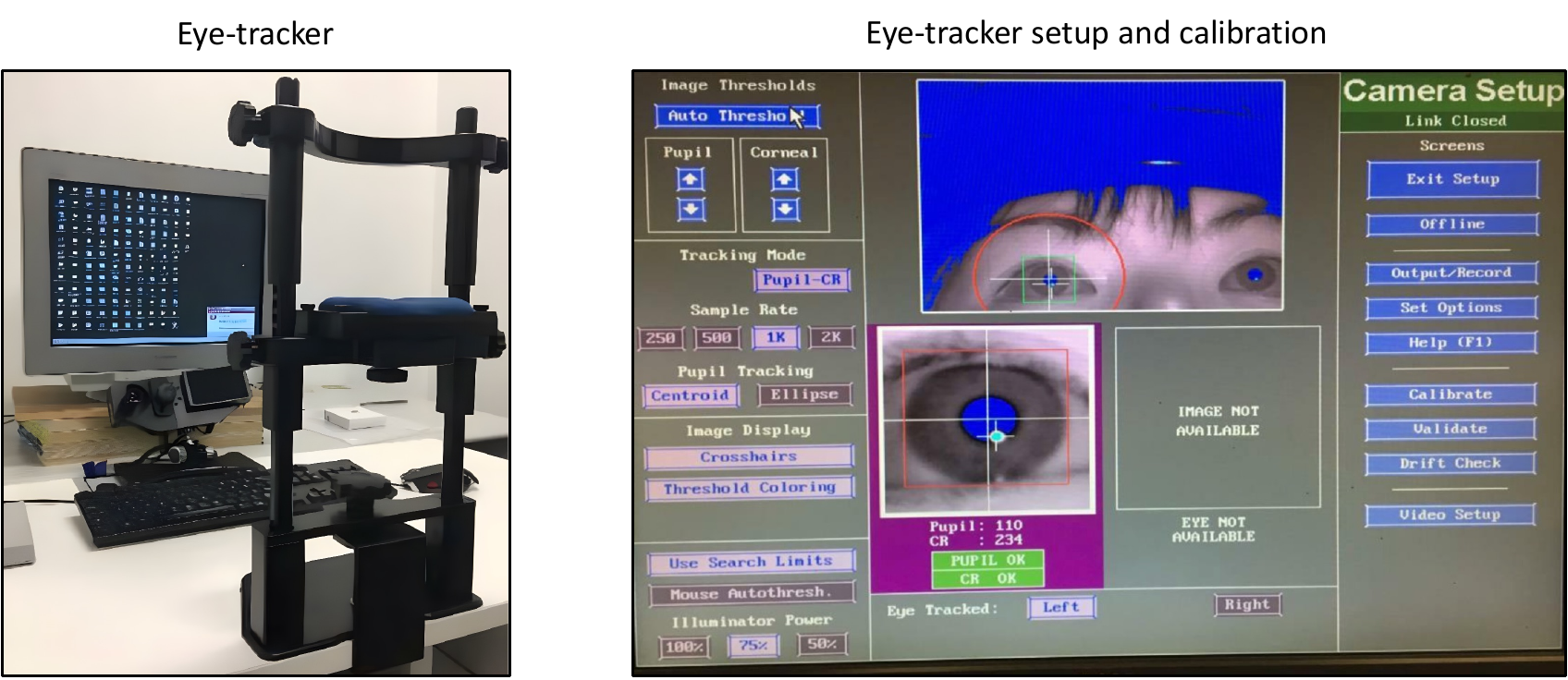}
    \caption{Eye-tracking equipment and setup. We use a chin rest to stabilize the eye-tracking process. During gaze collection, only one image is displayed on the screen. Annotators use their eyes to annotate and their hands to press a key to switch to the next image once they have finished.}
\end{figure}

\newpage

\section*{B. Implementation Details}

\begin{table}[h!]
    \centering
    \caption{Hyper-parameters used in experiments on two datasets.}
    {\renewcommand{\arraystretch}{1.2}\begin{tabular}{l@{\hspace{8pt}}|@{\hspace{8pt}}c@{\hspace{8pt}}}
\hline\hline

Backbone & 2D UNet\\
Supervision Loss & Cross-entropy loss \\
Training Iterations & 15000\\
Batch Size & 8\\
Optimizer & SGD \\
SGD Momentum $\mu$ & 0.99 \\
Scheduler & CosineAnnealingLR\\
Base Learning Rate & $1e^{-2}$ \\
Minimum Learning Rate & $1e^{-4}$ \\
\hline
Resolution  &  $224\times 224$ \\
Data Augmentation  & Random Flip \\
\multirow{2}{*}{Data Split} &  Kvasir-SEG: 900 training and 100 testing images\\
 & NCI-ISBI: 789 training and 117 testing images\\

\hline\hline
\end{tabular}}
    \label{tab:implementation}
\end{table}

\clearpage

\end{document}